\documentclass[11pt]{article}

\usepackage[utf8]{inputenc}
\usepackage{amsmath, amssymb, amsthm, mathtools, bbm, graphicx, fullpage, caption, subcaption, authblk, array, multirow, makecell, xurl, longtable, hyperref}
\usepackage[dvipsnames]{xcolor}
\usepackage[normalem]{ulem}
\usepackage{setspace}
\newcommand*{\abs}[1]{{\left|{#1}\right|}}

\title{Voting of predictive models for clinical outcomes: consensus of algorithms for the early prediction of sepsis from clinical data and an analysis of the PhysioNet/Computing in Cardiology Challenge 2019}

\author[1]{Matthew A. Reyna}
\author[1, 2]{Gari D. Clifford}
\affil[1]{\small{Department of Biomedical Informatics, Emory University}}
\affil[2]{\small{Department of Biomedical Engineering, Georgia Institute of Technology and Emory University}}
\date{December 20, 2020}

\begin{document}

\maketitle

%
%

\begin{abstract}
Although there has been significant research in boosting of weak learners, there has been little work in the field of boosting from strong learners. This latter paradigm is a form of weighted voting with learned weights. In this work, we consider the problem of constructing an ensemble algorithm from 70 individual algorithms for the early prediction of sepsis from clinical data. We find that this ensemble algorithm outperforms separate algorithms, especially on a hidden test set on which most algorithms failed to generalize.
\end{abstract}

%
%

\section{Introduction}

By posing clinical problems as prediction or classification tasks, researchers can train computational models on routinely available clinical data to solve clinically relevant problems. Recently, such computational tools have been shown to perform as well or better than domain experts at several important clinical tasks (see \cite{Poplin2018, Nemati2018, Cheng2019} among many others).

However, the promise of artificial intelligence (AI) and big data remains largely unrealized in healthcare. This unrealized potential has many causes, including general issues in AI as well as other issues that are more specific to health and healthcare \cite{Clifford2020, Leisman2020}. The PhysioNet/Computing in Cardiology (CinC) Challenges address many of these issues by posing clear problem definitions, sharing well characterized and curated databases from diverse geographical locations, and defining evaluation metrics for algorithms that capture the importance of the algorithms in a clinical setting \cite{Goldberger2000}. Jointly hosted by PhysioNet and CinC, these annual Challenges have addressed clinically interesting questions that are unsolved or not well solved for over twenty years.

The PhysioNet/CinC Challenge 2019, hereafter described as either the Challenge or the 2019 Challenge, asked participants to design algorithms for the early prediction of sepsis from routinely available clinical data \cite{PhysioNet2019}. For the Challenge, we curated electronic medical records (EMRs) for over 60,000 ICU patients from three distinct hospital systems. These records had up to 40 clinical variables for each hour of the patient's ICU stay. We also introduced a novel, time-dependent evaluation metric to assess the clinical utility of the algorithms' predictions.

A total of 104 teams from academia and industry submitted 853 algorithms for evaluation in the Challenge, and 90 abstracts were accepted for presentation at CinC 2019. Each team was allowed to nominate one of their algorithms for evaluation on the hidden test data, resulting in 88 algorithms for early sepsis prediction\footnote{We were unable to score 16 algorithms on the full test dataset, so we do not consider them in this article.}. In this article, we focus on 70 algorithms that were most promising for further analysis\footnote{We were unable to score an additional 11 algorithms on the full training data, and we were able to score another 7 algorithms that performed no better than an inactive method that made only negative predictions on at least one of the training sets, so we do not focus on them in this article.}.

These algorithms represent a diversity of approaches to early sepsis prediction.
We ranked these algorithms based on their performance on the hidden test datasets using the clinically derived evaluation metric that we developed for the Challenge. However, while some algorithms necessarily performed better than others, many lower-ranked algorithms outperformed higher-ranked on certain examples.
In some cases, this specialization was the direct and desired result of feature engineering or other model design decisions, but in other cases, it was an unintended consequence of the way a model is constructed and implemented\footnote{We actively sought to preserve the diversity of the Challenge algorithms by prohibiting teams from collaborating.}. Indeed, previous Challenges found that simple voting models were able to outperform individual models for the classification of electrocardiograms (ECGs) and phonocardiograms (PCGs) \cite{PhysioNet2013, PhysioNet2016, PhysioNet2017}. This `wisdom of the crowd' applies more generally to computational approaches and clinical applications \cite{Hong2004, Han2019, Johnson2019}. 

In this article, we investigate the diversity of the 2019 Challenge algorithms and describe a simple voting model for the Challenge that outperforms that individual Challenge algorithms. The voting model's performance is especially important on a completely hidden test set, which allows the assessment of the ability of models to generalize to new databases \cite{Johnson2019, Clifford2020}.

%
%

\section{The PhysioNet/Computing in Cardiology Challenge 2019}

The PhysioNet/Computing in Cardiology Challenge 2019, again described as either the Challenge or the 2019 Challenge, introduced curated and labeled datasets, a problem statement, and a novel evaluation metric to facilitate the development of diverse computational approaches for early sepsis prediction. We provide a more complete description in the reference paper for the 2019 Challenge \cite{PhysioNet2019}.

\subsection{Challenge Data}

We curated and labeled electronic medical records (EMRs) for over 63,097 ICU patient encounters from three distinct hospital systems. A total of 40,336 records from two hospital systems were shared as public training sets, and the remaining 22,761 records from three hospital systems were sequestered as hidden test sets. In particular, one hospital system was used only for test data, allowing us to assess how algorithms generalized to new hospital systems.

\begin{figure}
    \centering
    \includegraphics[scale=0.8]{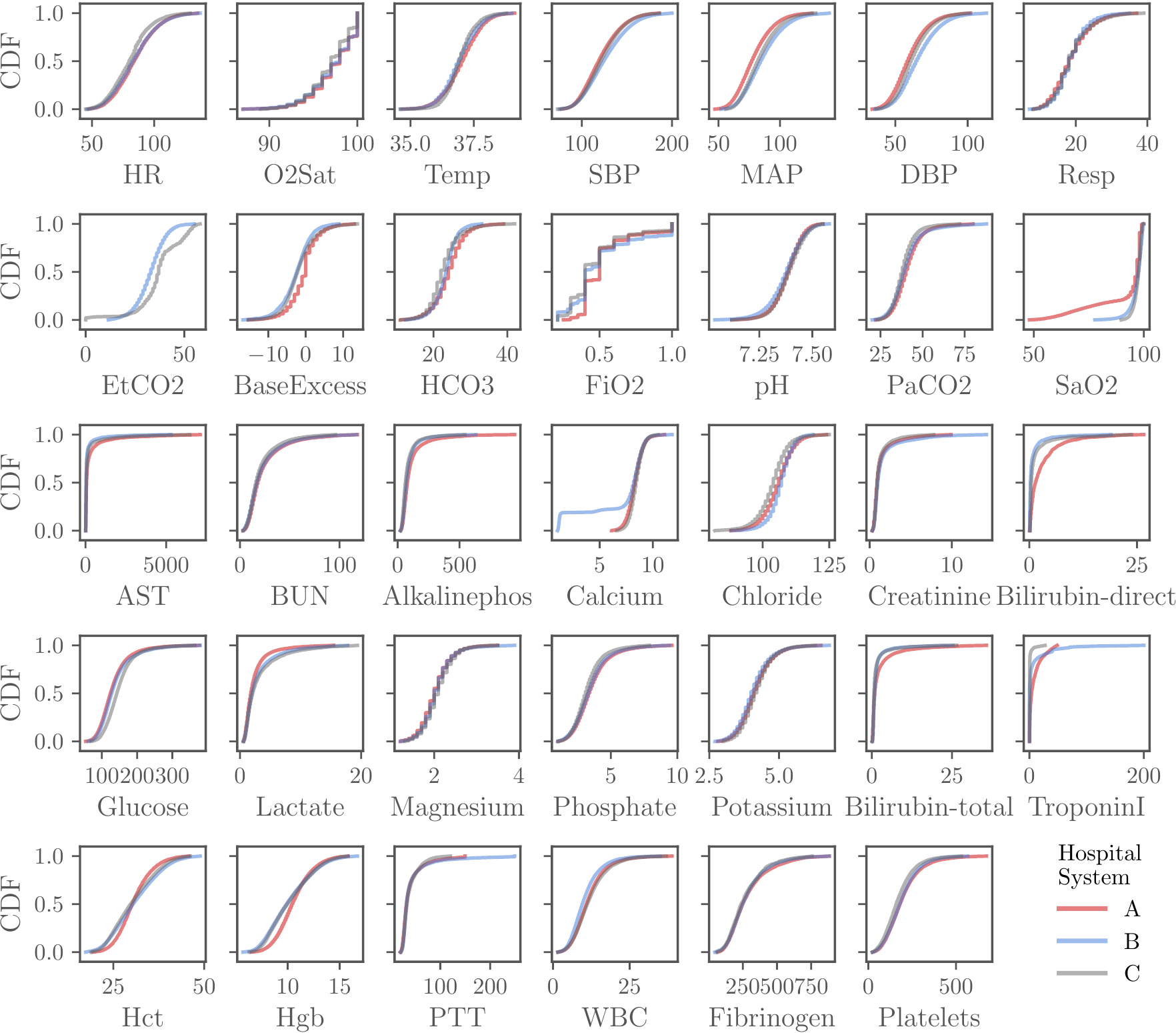}
    \caption{Cumulative distribution functions (CDFs) of the clinical variables provided in the Challenge data across different institutions; see Table \ref{table:clinical_data} for the variable abbreviations and units.}
    \label{fig:cdfs}
\end{figure}

These records contained up to 40 different routinely collected clinical variables or measurements for each hour of a patient's ICU stay.
Some variables, such as heart rate or temperature, were taken multiple times each hour and summarized by the median value of the measurements over the course of the hour. Other variables, such as lactate, were taken less frequently or irregularly and were missing in most hourly time windows. Still other variables, such as age and sex, were generally available and remained constant throughout a patient's stay. Table \ref{table:clinical_data} provides a full list of the available clinical variables, and Figure \ref{fig:cdfs} shows how the distributions of these variables differed across datasets, reflecting differences in patient populations and/or clinical practices at different institutions.

\begin{table}[tbp]
\centering
{\footnotesize
\begin{tabular}{|r|l|l|} \hline
  & Variable Abbreviation & Variable Description (units) \\ \hline
1 & \texttt{HR} & Heart rate (beats per minute)\\ \hline
2 & \texttt{O2Sat} & Pulse oximetry (\%)\\ \hline
3 & \texttt{Temp} & Temperature (deg C)\\ \hline
4 & \texttt{SBP} & Systolic BP (mm Hg)\\ \hline
5 & \texttt{MAP} & Mean arterial pressure (mm Hg)\\ \hline
6 & \texttt{DBP} & Diastolic BP (mm Hg)\\ \hline
7 & \texttt{Resp} & Respiration rate (breaths per minute)\\ \hline
8 & \texttt{EtCO2} & End tidal carbon dioxide (mm Hg)\\ \hline
9 & \texttt{BaseExcess} & Excess bicarbonate (mmol/L)\\ \hline
10 & \texttt{HCO3} & Bicarbonate (mmol/L)\\ \hline
11 & \texttt{FiO2} & Fraction of inspired oxygen (\%)\\ \hline
12 & \texttt{pH} & pH\\ \hline
13 & \texttt{PaCO2} & Partial pressure of carbon dioxide from arterial blood (mm Hg)\\ \hline
14 & \texttt{SaO2} & Oxygen saturation from arterial blood (\%)\\ \hline
15 & \texttt{AST} & Aspartate transaminase (IU/L)\\ \hline
16 & \texttt{BUN} & Blood urea nitrogen (mg/dL)\\ \hline
17 & \texttt{Alkalinephos} & Alkaline phosphatase (IU/L)\\ \hline
18 & \texttt{Calcium} & Calcium (mg/dL)\\ \hline
19 & \texttt{Chloride} & Chloride (mmol/L)\\ \hline
20 & \texttt{Creatinine} & Creatinine (mg/dL)\\ \hline
21 & \texttt{Bilirubin-direct} & Direct bilirubin (mg/dL)\\ \hline
22 & \texttt{Glucose} & Serum glucose (mg/dL)\\ \hline
23 & \texttt{Lactate} & Lactic acid (mg/dL)\\ \hline
24 & \texttt{Magnesium} & Magnesium (mmol/dL)\\ \hline
25 & \texttt{Phosphate} & Phosphate (mg/dL)\\ \hline
26 & \texttt{Potassium} & Potassium (mmol/L)\\ \hline
27 & \texttt{Bilirubin-total} & Total bilirubin (mg/dL)\\ \hline
28 & \texttt{TroponinI} & Troponin I (ng/mL)\\ \hline
29 & \texttt{Hct} & Hematocrit (\%)\\ \hline
30 & \texttt{Hgb} & Hemoglobin (g/dL)\\ \hline
31 & \texttt{PTT} & Partial thromboplastin time (seconds)\\ \hline
32 & \texttt{WBC} & Leukocyte count (count/L)\\ \hline
33 & \texttt{Fibrinogen} & Fibrinogen concentration (mg/dL)\\ \hline
34 & \texttt{Platelets} & Platelet count (count/mL)\\ \hline
35 & \texttt{Age} & Age (years) \\ \hline
36 & \texttt{Gender} & Female ($0$) or male ($1$)\\ \hline
37 & \texttt{Unit1} & Identifier for medical ICU unit; false ($0$) or true ($1$)\\ \hline
38 & \texttt{Unit2} & Identifier for surgical ICU unit; false ($0$) or true ($1$)\\ \hline
39 & \texttt{HospAdmTime} & Time between hospital and ICU admission (hours)\\ \hline
40 & \texttt{ICULOS} & ICU length of stay (hours)\\ \hline
\end{tabular}
}
\caption{Clinical variables used for the Challenge, including 8 vital sign values, 28 laboratory values, and 6 demographic values.}
\label{table:clinical_data}
\end{table}

Each hourly time window of each patient record was labeled according to Sepsis-3 clinical criteria for sepsis onset \cite{Seymour2016, Singer2016, Shankar2016}. For each patient that eventually satisfied these criteria, we specified the following three time points to define the onset time $t_\text{sepsis}$ of sepsis:
\begin{itemize}
    \item $t_\text{suspicion}$:
    Clinical suspicion of infection was identified as the earlier timestamp of intravenous (IV) antibiotics and blood cultures within a given time interval. If IV antibiotics were given before cultures, then the cultures must have been obtained within 24 hours. If cultures were obtained before IV antibiotics, then IV antibiotic must have been ordered within 72 hours. In either case, IV antibiotics must have been administered for at least 72 consecutive hours.
    \item $t_\text{SOFA}$:
    Organ failure was identified by a two-point increase in the Sequential Organ Failure Assessment (SOFA) score within a 24-hour period.
    \item $t_\text{sepsis}$: The onset of sepsis was identified as the earlier of $t_\text{suspicion}$ and $t_\text{SOFA}$ as long as $t_\text{SOFA}$ occurred no more than 24 hours before or 12 hours after $t_\text{suspicion}$.
\end{itemize}

We intentionally and explicitly preserved missing and erroneous data as part of the Challenge. However, we did not include patients with less than 8 hourly time windows of data in the ICU or patients with $t_\text{sepsis}$ less than 4 hours after ICU admission. We also truncated patient records after ICU discharge, and we truncated patient records with more than two weeks of measurements to only two weeks.

\subsection{Challenge Objective and Evaluation}

The goal of the Challenge was to predict sepsis from the routinely collected clinical data.
We challenged participants to design algorithms for predicting sepsis six hours before clinical recognition of sepsis onset.
Each algorithm made an hourly prediction for whether the patient met Sepsis-3 criteria for sepsis onset \cite{Seymour2016, Singer2016, Shankar2016}. We assigned a numerical score to each hourly prediction in each patient record based on the relative clinical utility of the prediction. Correct sepsis predictions were rewarded up to 12 hours before sepsis onset, false alarms were lightly penalized, and failures to detect sepsis near sepsis onset were strongly penalized. We motivate and describe this scoring system in greater detail in \cite{PhysioNet2019}.

\subsection{Challenge Algorithms}

By the end of the Challenge, 104 teams from academia and industry had submitted 853 sepsis prediction algorithms for evaluation in the Challenge. Teams were required to submit their algorithms as containerized entries to improve the reproducibility and usability of their code. We evaluated each of these entries on hidden test data in a cloud environment, and teams with one or more successful entries (e.g., did not crash, did not exceed resource limits, etc.) were able to nominate one of their successful entries for evaluation on the full test data. The teams with the highest clinical utility score on the full test data won the Challenge.

Table \ref{table:results} outlines the highest scoring teams in the Challenge, and Table \ref{table:teams} provides more information about all 88 entries that we successfully evaluated on the test data. Most methods, including the highest performing ones, performed noticeably better on the test data from hospital systems A and B than on the test data from hospital system C, reflecting the availability of training data from hospital systems A and B but not from hospital system C.

\begin{table}[tbp]
\centering
{\footnotesize
\begin{tabular}{cp{2cm}p{6cm}p{1.2cm}p{1cm}p{1cm}p{1cm}}
    \hline
    Rank & Team Name & Team Members & Full Test Set & Test Set A & Test Set B & Test Set C \\\hline
    1 & Can I get your signature? & James Morrill, Andrey Kormilitzin, Alejo Nevado-Holgado, Sumanth Swaminathan, Sam Howison, Terry Lyons & 0.360 & 0.433 & 0.434 & -0.123 \\
    2 & Sepsyd & John Anda Du, Nadi Sadr, Philip de Chazal & 0.345 & 0.409 & 0.396 & -0.042 \\
    3 & Separatrix & Morteza Zabihi, Serkan Kiranyaz, Moncef Gabbouj & 0.339 & 0.422 & 0.395 & -0.146 \\
    4 & FlyingBubble & Xiang Li, Yanni Kang, Xiaoyu Jia, Junmei Wang, Guotong Xie	& 0.337 & 0.420 & 0.401 & -0.156 \\
    5 & CTL-Team & Janmajay Singh, Kentaro Oshiro, Raghava Krishnan, Masahiro Sato, Tomoko Ohkuma, Noriji Kato & 0.337 & 0.401 & 0.407 & -0.094 \\ \hline
    * & SailOcean & Meicheng Yang, Hongxiang Gao, Xingyao Wang, Yuwen Li, Jianqing Li, Chengyu Liu & 0.364 & 0.430 & 0.422 & -0.048
\end{tabular}
}
\caption{Clinical utility scores for the teams with the five highest scores on the full test set from hospital systems A, B, and C, i.e., as well as their scores on the separate test sets from hospital systems A, B, and C. * denotes the highest-scoring unofficial entry that did not meet the full Challenge requirements.}
\label{table:results}
\end{table}

We were able to successfully run entries from 88 teams on the test data, resulting in 88 successful algorithms; we were unable to run code from 16 teams on the test data due to coding errors, exceeded resource limits, or non-compliance with Challenge rules. Of these 88 entries, we were able to successfully run 77 algorithms on the training data; we were unable to run code from 11 teams on the training data due to additional coding errors or exceeded resource limits. Of these 77 entries, we focus on 70 algorithms that received positives utility scores on both of the training sets; by definition, entries with non-positive utility scores performed no better than an inactive method that only reported negative sepsis predictions. These 70 entries formed the basis of our analysis.

These entries include MATLAB, Python, and R implementations of a diversity of algorithms ranging from simple regression models with engineered features to complex deep learning models that utilize various libraries, including TensorFlow \cite{TensorFlow}, PyTorch \cite{PyTorch}, Keras \cite{Keras}, and XGBoost \cite{XGboost}. These differences contribute to the diversity of the Challenge entries.

\section{Challenge Voting}

We explore the diversity of the Challenge models and construct a simple voting model that outperforms the individual models on the Challenge test data.

\subsection{Model Diversity}

Previous Challenges found that simple voting models outperformed individual models for the classification of electrocardiograms (ECGs) and phonocardiograms (PCGs) \cite{PhysioNet2013, PhysioNet2016, PhysioNet2017}, demonstrating the ``wisdom of the crowd'' for computational approaches for clinical tasks \cite{Hong2004, Han2019, Johnson2019}. However, these analyses did not explicitly consider the diversity of these approaches that makes ensemble or voting approaches successful. 

To safeguard the diversity of the Challenge entries, we imposed rules against collaboration and employed several features to check for collaboration and plagiarism. Indeed, we detected the contravention of these rules in several cases because of highly similar code bases, predictions, or other irregularities, resulting in the disqualification of multiple teams; none of the disqualified approaches appear in this article. Indeed, many of the comparisons of code bases are designed to detect plagiarism, e.g., the winnowing approach by the Measure of Software Similarity (Moss) \cite{Schleimer2003}, but these approaches focus on detecting identical or high similar code bases instead of quantifying the amount of similarity or dissimilarity between different code bases.

\begin{figure}[tbp]
\centering
\includegraphics[scale=1]{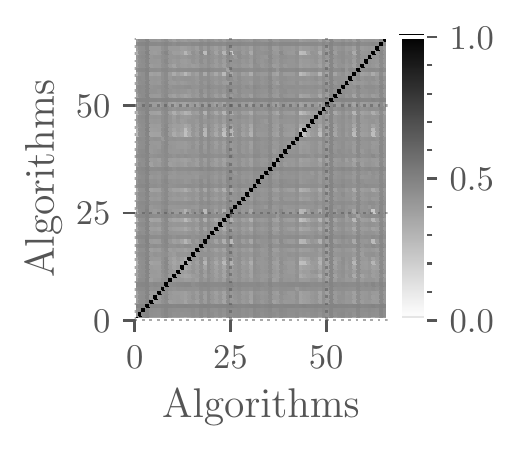}
\caption{Code similarity between algorithms. Rows and columns are 67 Python entries (sorted from highest to lowest utility scores on the full test set), and each entry is the pairwise similarity measure defined by the tree edit distance between each pair of abstract syntax trees for each method.}
\label{fig:code_similarity}
\end{figure}

First, we compared the Challenge algorithms by directly comparing their code bases. For this initial comparison, we compute the abstract syntax tree (AST) for each Python code base and define a pairwise (dis)similarity measure\footnote{For visualization, we convert this distance or dissimilarity measure $d$ to a similarity measure $s$ by computing $s = 1/d$.} between two code bases using the tree edit distance between their respective ASTs \cite{Zhang1989}.  Figure \ref{fig:code_similarity} shows the pairwise similarity between each pair of algorithms from this comparison.

Unfortunately, this approach is limited. It can only compare comparable code bases; it is more difficult to compare codes from different programming languages or even from different versions of the same programming language, e.g., Python 2.x with Python 3.x. More importantly, it does not reflect the model parameters, which are not part of the code's AST, so wildly different deep learning approaches may have high similarity. It also does not fully reflect the differences between models on different datasets because the code is the same for each dataset.

Next, we compared the Challenge algorithms by comparing their predictions. For this comparison, we use the Jaccard index between the predictions from every pair of Challenge entries. For algorithms $i$ and $j$ with sequences of hourly binary predictions $(x_k)_k$ and $(y_k)_k$, respectively, we compute the pairwise similarity score
\begin{equation}
    \label{eq:unweighted-score}
    \sigma_{ij} = \frac{\sum_{k} x_k y_k}{\sum_{k} x_k \lor y_k}.
\end{equation}

\begin{figure}[tbp]
\centering
\begin{subfigure}[t]{0.19\textwidth}
\centering
\includegraphics[scale=0.55]{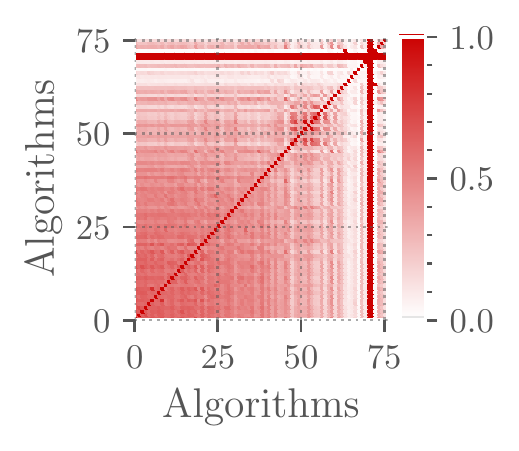}
\caption{Training set A}
\end{subfigure}
\hfill
\begin{subfigure}[t]{0.19\textwidth}
\centering
\includegraphics[scale=0.55]{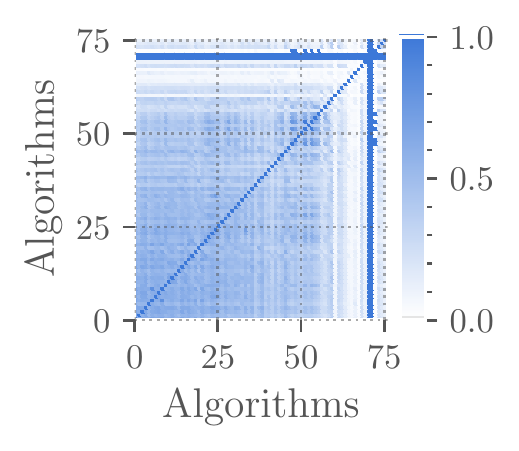}
\caption{Training set B}
\end{subfigure}
\hfill
\begin{subfigure}[t]{0.19\textwidth}
\includegraphics[scale=0.55]{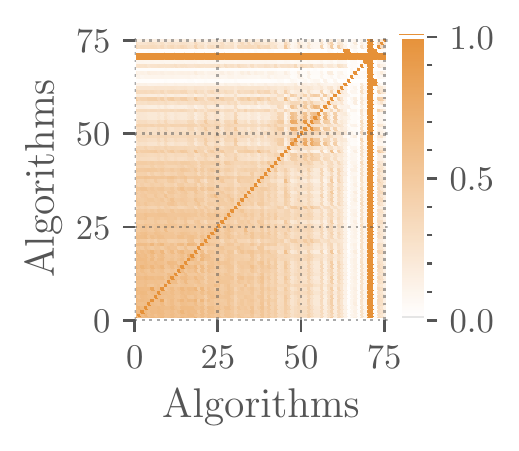}
\caption{Test set A}
\end{subfigure}
\hfill
\begin{subfigure}[t]{0.19\textwidth}
\includegraphics[scale=0.55]{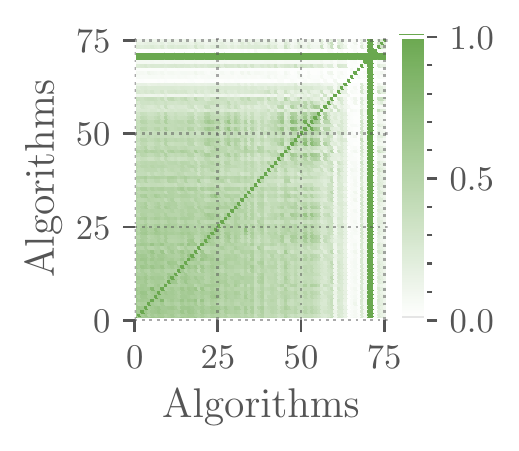}
\caption{Test set B}
\end{subfigure}
\hfill
\begin{subfigure}[t]{0.19\textwidth}
\includegraphics[scale=0.55]{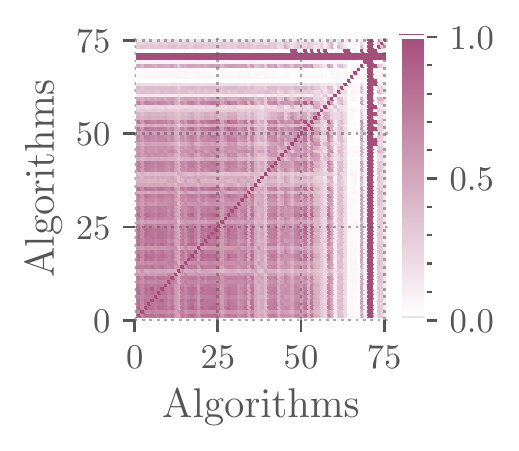}
\caption{Test set C}
\end{subfigure}
\caption{Unweighted prediction similarity between algorithms on the training and test sets from hospital systems A, B, and C. Rows and columns are 77 entries (sorted from highest to lowest utility scores on the full test set), and each entry is the pairwise similarity measure defined by \eqref{eq:weighted-score}.}
\label{fig:unweighted_prediction_similarity}
\end{figure}

Figure \ref{fig:unweighted_prediction_similarity} shows the pairwise similarity between each pair of algorithms from this comparison on each of the training and test sets. With this comparison, the algorithms demonstrate different degrees of similarity across datasets with especially high similarity on the test data for hospital system C, which was not represented in the training data; the high similarity on test set C reflects consistent errors by many methods on this hospital system. Moreover, the highest performing algorithms tend to be more similar, but several lower performing algorithms also have high similarity, and several other algorithms appear to be completely dissimilar from the others.

This approach to comparison is also limited. While it reflects differences between models on different datasets, it weighs each prediction equally, so differences between model predictions for clinically important predictions are given the same value as unimportant ones. For example, while some models have have different patterns of false alarms, the most important predictions for sepsis are those for patients in the several hours before or after sepsis onset.

Finally, we compared the Challenge algorithms by comparing their predictions while using our evaluation metric to assign more weight to predictions that are more clinically important.
For this comparison, we define a weighted Jaccard-like index between the predictions from every pair of Challenge entries. For algorithms $i$ and $j$ with a sequence of utility score values $(u_k)_k$ for a sequence of hourly binary predictions $(x_k)_k$ and a sequence of utility score values $(v_k)_k$ for a sequences of hourly binary predictions $(y_k)_k$, respectively, we compute the weighted pairwise similarity score
\begin{equation}
    \label{eq:weighted-score}
    \sigma_{ij}' = 1 - \frac{\sum_{k} \abs{u_k - v_k}}{\sum_{k} \abs{u_k} + \abs{v_k}}.
\end{equation}

Figure \ref{fig:unweighted_prediction_similarity} shows the pairwise similarity between each pair of algorithms from this comparison on each of the training and test sets. While these weighted similarity scores may be expected to be distributed differently from the unweighted pairwise similarity scores, the highest scoring models demonstrate qualitatively more similarity with one another and more dissimilarity with the lowest scoring models.

\begin{figure}[tbp]
\centering
\begin{subfigure}[t]{0.19\textwidth}
\centering
\includegraphics[scale=0.5]{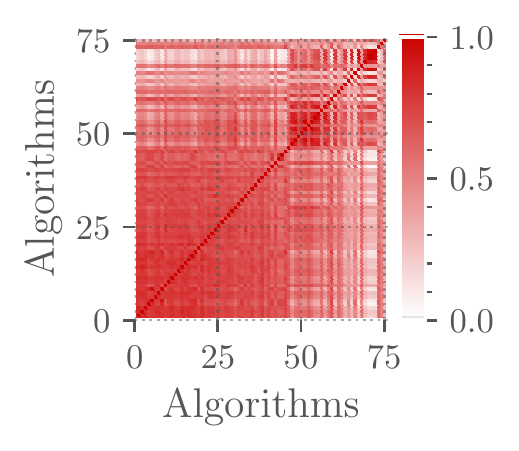}
\caption{Training set A}
\end{subfigure}
\hfill
\begin{subfigure}[t]{0.19\textwidth}
\centering
\includegraphics[scale=0.5]{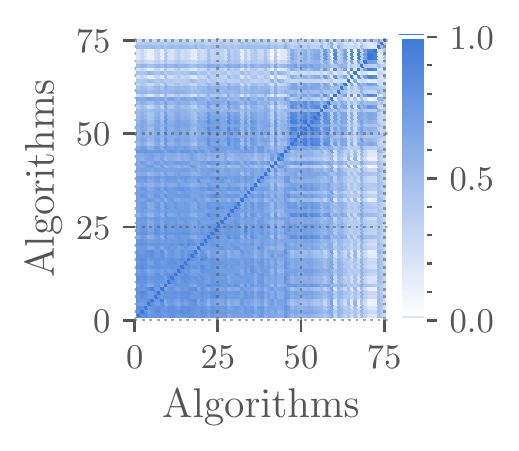}
\caption{Training set B}
\end{subfigure}
\hfill
\begin{subfigure}[t]{0.19\textwidth}
\includegraphics[scale=0.5]{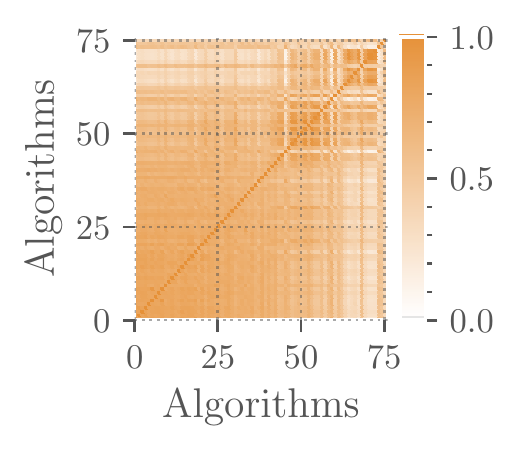}
\caption{Test set A}
\end{subfigure}
\hfill
\begin{subfigure}[t]{0.19\textwidth}
\includegraphics[scale=0.55]{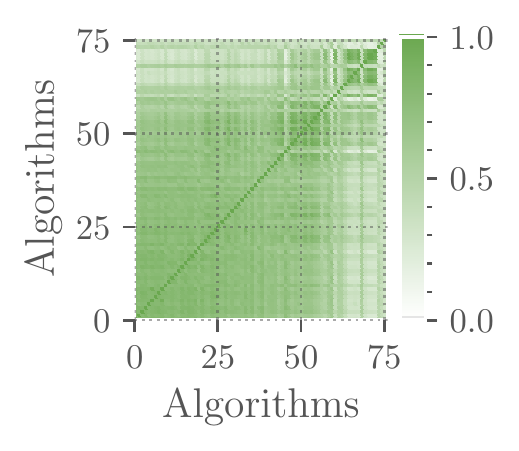}
\caption{Test set B}
\end{subfigure}
\hfill
\begin{subfigure}[t]{0.19\textwidth}
\includegraphics[scale=0.55]{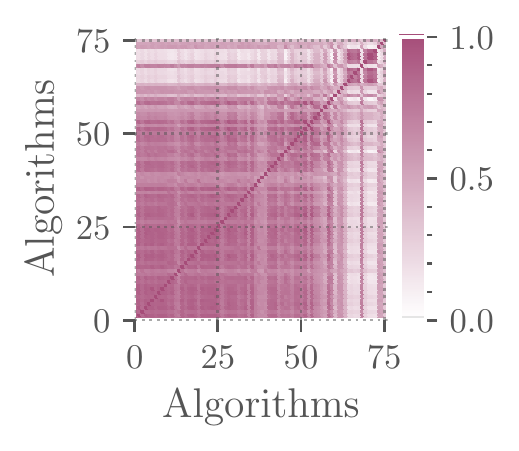}
\caption{Test set C}
\end{subfigure}
\caption{Weighted prediction similarity between algorithms on the training and test sets from hospital systems A, B, and C. Rows and columns are 77 entries (sorted from highest to lowest utility scores on the full test set), and each entry is the pairwise similarity measure defined by \eqref{eq:weighted-score}.}
\label{fig:weighted_prediction_similarity}
\end{figure}

\begin{figure}[tbp]
\centering
\begin{subfigure}[t]{0.325\textwidth}
\includegraphics[scale=0.9]{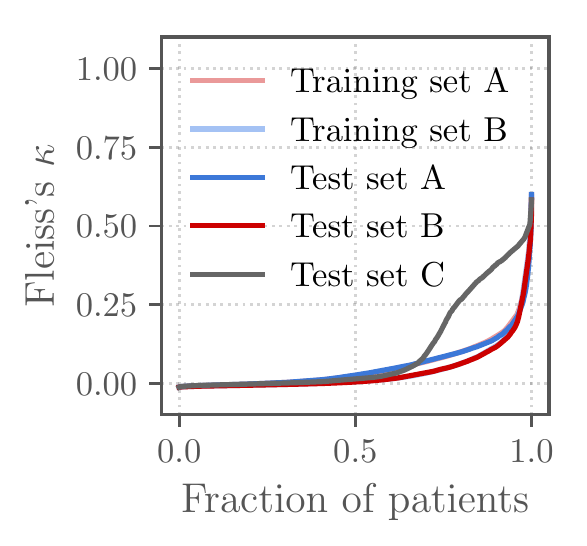}
\caption{All 77 models}
\end{subfigure}
\hfill
\begin{subfigure}[t]{0.325\textwidth}
\includegraphics[scale=0.9]{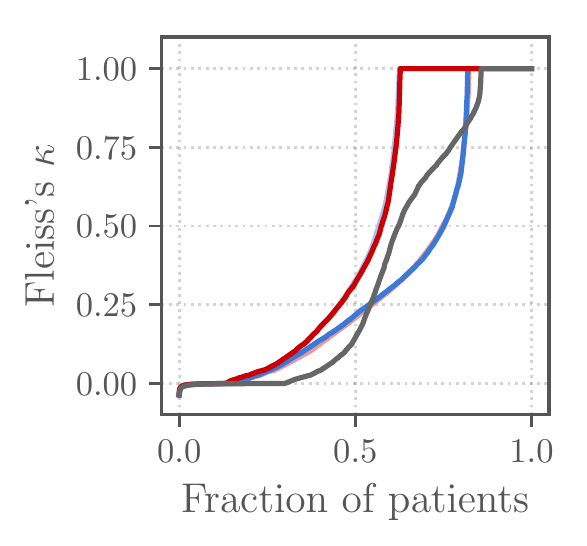}
\caption{Top-scoring 25 models}
\end{subfigure}
\hfill
\begin{subfigure}[t]{0.325\textwidth}
\includegraphics[scale=0.9]{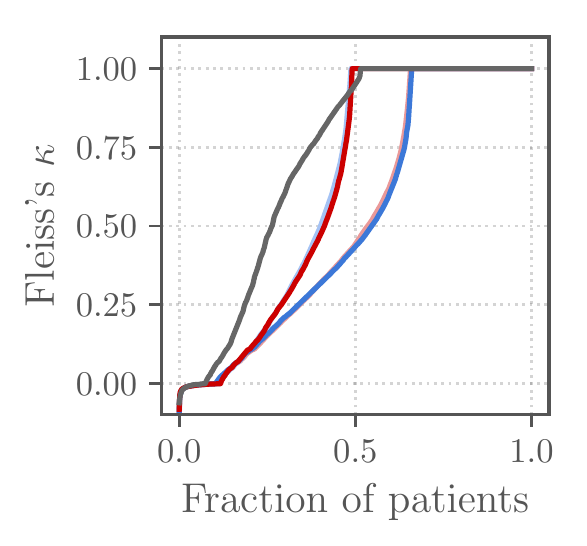}
\caption{Top-scoring 10 models}
\end{subfigure}
\caption{Fleiss' $\kappa$ for patients in hospitals A, B, and C for all models, the top 25 models, and the top 10 models by utility score on the full test set.}
\label{fig:fleiss}
\end{figure}

Our observations about high similarity between high-scoring models are further by the agreement between these models on specific patients. Figure \ref{fig:fleiss} directly demonstrates high inter-rater reliability between the highest scoring models on a majority of patient records and low inter-rater reliability between the larger set of models on all but a small fraction of patients. The high inter-rater reliability for the highest scoring models on the test sets for hospital systems B and C (but not A) are particularly interesting because these models performed considerably better on the test sets from hospital systems A and B than they did on the test set from hospital system C, which was not represented in the training data.

\subsection{Voting Algorithm}

We devised a simple consensus voting algorithm to combine individual algorithms. For each hourly time window in a patient record, this algorithm estimates the confidence in the voting model using the amount of agreement or disagreement between the individual algorithms' predictions at each time point.

We employed a greedy selection procedure (with replacement, so a model may be chosen multiple times and contribute a higher weight to the voting model) on the training data from hospital systems A and B. We stopped adding additional models when the greedy selection procedure stopped improving the score of the model on the training data. 
We used this process to train two models: one model for hospital systems A and B and the other model for hospital system C.

First, since the training data contained patient records from hospital systems A and B, we expected that the individual models would perform reasonably well on the test data from hospital systems A and B. Accordingly, we used majority votes for our first voting model, i.e., if a majority of models made positive predictions, then the voting model also made a positive prediction, and if a majority of models made negative predictions, then the voting model made a negative prediction\footnote{For ties, the voting model also made a positive prediction.}.

Second, since the training data did not contain patient records from hospital system C, we expected that the individual models would not perform as well on hospital system C. Therefore, due to the high concordance between model predictions on the test set from hospital system C, we required a more conservative approach than majority votes and required that all but one model make a positive prediction for the voting model to make a positive prediction.

\begin{figure}
    \centering
    \includegraphics[scale=0.9]{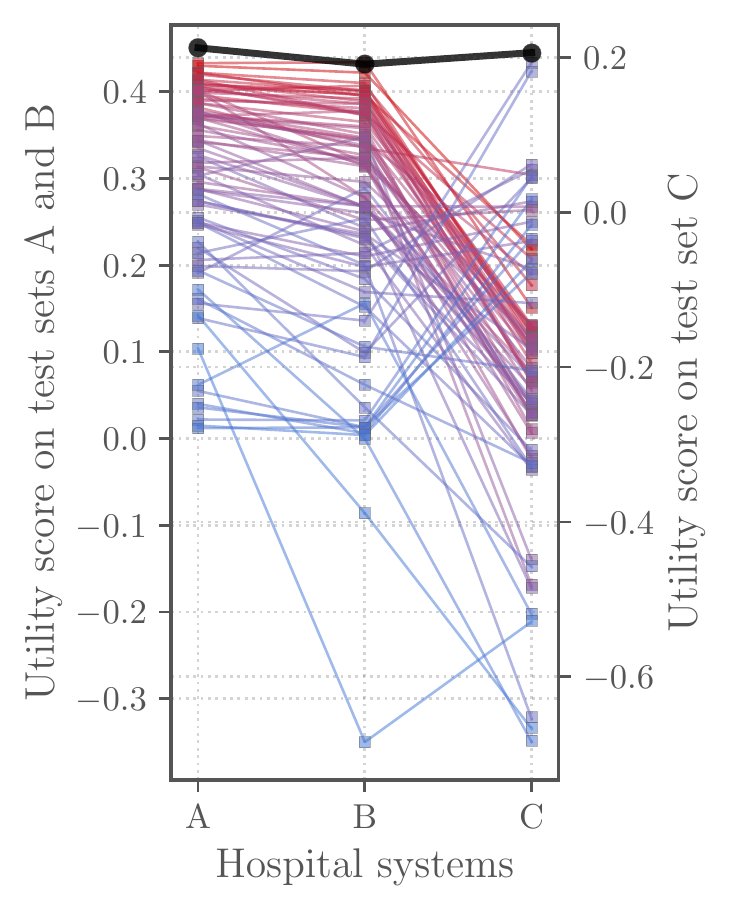}
    \caption{Clinical utility scores of 70 individual algorithms and the consensus voting algorithm on the test data for hospital systems A, B, and C. Individual algorithms are shown with colored squares and lines, where red indicates the highest score on the full test set and blue indicates the lowest score on the full test set. The voting algorithm is shown with black circles and lines.}
    \label{fig:ranks}
\end{figure}

This consensus voting model switches between these two different voting models depending on the agreement between model predictions on the test data and expectations of good or poor generalizability on the test data. Figure \ref{fig:ranks} shows that the voting model modestly outperforms the individual models on each of the test sets.

It is important to note that the voting model was not influenced by the clinical utility scores on the test data. Instead, we used the scores on the training data, the similarity between methods on the training and test data, and expectations about model performance on a completely hidden dataset\footnote{Of course, a fully automated approach that leverages these observations would be preferable.}.

%
%

\section{Discussion}

We show that ensemble models can outperform individual models for early sepsis predictions. Earlier analyses demonstrated the potential of voting models for clinical classification tasks \cite{PhysioNet2013, PhysioNet2017, PhysioNet2019}, and this simple approach continues to demonstrate the potential of voting models for clinical prediction tasks.

The diversity of the approaches used for the voting model determines the potential of the voting model. If methods are highly similar, then any voting model that is defined from them is likely to be highly similar as well, limiting the potential for improvement. While we believe that there are opportunities for improvement for this simple voting model, the high concordance between the high performing methods is responsible for a large share of the modest performance improvements of this voting model over the individual models. Moreover, while the poor generalizability of the individual models on the test set from hospital system C provides an opportunity for improvement, the high agreement between models on this data further limits that opportunity. In other words, if these models generalized poorly but in different ways, then there would be more opportunity for improvement than if they generalized poorly in the same way; unfortunately, the latter was the case.

Another opportunity for diversity lies not just in the trained models but in training the models. For example, it may be beneficial to develop an end-to-end voting model that could retrain the individual models and allow them to specialize on subpopulations in the data, increasing the diversity of the resulting models. In related work \cite{Nasiri2020}, we demonstrated that clustering subpopulations, training models on each population, and then weighting models by an individual's distance (in parameter space) from each cluster, we substantially improved the performance of an algorithm. This `semi-personalized' approach to modeling could be improved by selecting independent algorithms that perform better on a particular subpopulation. Both of these approaches introduces substantially more complexity and computational demands during training, but little extra work during the forward use of the models. 
Finally, we note that the parameters of the cost function that we proposed for the 2020 PhysioNet/CinC Challenge could also be optimized at the same time as the predictors.

There are a two key limitations of the analysis we have provided in this work. First, there is no way to prove definitively that test set C was comparable to training and test sets A and B. Although we matched the data in a univariate statistical manner, covariates in both space and time may be significantly different, due to subtle but important differences in the patient population and clinical practices. It may be that test set C was so significantly different (in a covariate space) that without providing data from test set C, no algorithm could be expected to generalize to that test set. (We note that some weaker algorithms did managed to do so, but those models are essentially inadequate in general.) Second, although groups were required to work without collaboration, we cannot guarantee that some groups (particularly those that were of most importance in the weighted voting), did not engage in collusion. We deem this highly unlikely, given our analysis of the submissions. However, there is a likelihood that the proliferation of standard libraries for machine learning have created significant relationships between code bases which break the independence assumption. Future work should investigate methods to better measure independence based on code structure, not algorithm outputs.    

%
%

\section*{Acknowledgements}

This work was funded by the Gordon and Betty Moore Foundation, the National Science Foundation, grant \# 1822378  `Leveraging Heterogeneous Data Across International Borders in a Privacy Preserving Manner for Clinical Deep Learning', the National Institute of General Medical Sciences (NIGMS) and the National Institute of Biomedical Imaging and Bioengineering (NIBIB) under NIH grant number 2R01GM104987-09 and the National Center for Advancing Translational Sciences of the National Institutes of Health under Award Number UL1TR002378. The content is solely the responsibility of the authors and does not necessarily represent the official views of the National Institutes of Health or National Science Foundation. Disclosures: GC holds equity and serves as Chief Science Officer of LifeBell AI, which produces products loosely related to the work presented in this article.

%
%

\bibliographystyle{unsrt}
\bibliography{references}

\newpage

%
%

\section*{Appendix}

This table describes all 88 entries that were successful on the full test set.

{\tiny\begin{longtable}{|l|l|l|l|r|r|r|r|r|r|}\hline
                     \makecell[l]{Team Name}  &  \makecell[l]{Language}  &  \makecell[l]{GPU?}  &  \makecell[l]{Focus of \\Analysis?} & \makecell[l]{Training\\Set A}  &  \makecell[l]{Training\\Set B}  &  \makecell[l]{Test\\Set A}  &  \makecell[l]{Test \\Set B}  &  \makecell[l]{Test\\Set C}  & \makecell[l]{Full\\Test Set} \\\hline 
    404: Sepsis not found  &    Python  &  False  &       True  &           0.415  &           0.370  &       0.369  &       0.346  &      -0.172  &          0.290  \\ 
                      ABC  &    Python  &  False  &       True  &           0.405  &           0.304  &       0.363  &       0.278  &      -0.314  &          0.247  \\ 
                AI4Sepsis  &    Python  &  False  &       True  &           0.378  &           0.326  &       0.344  &       0.314  &      -0.258  &          0.255  \\ 
           AI-Neuroimmune  &    Python  &  False  &       True  &           0.341  &           0.251  &       0.325  &       0.243  &      -0.485  &          0.192  \\ 
                  AlgTeam  &    MATLAB  &  False  &      False  &           0.405  &            None  &       0.335  &       0.268  &      -0.226  &          0.240  \\ 
        Amini-Univ-Tehran  &         R  &  False  &       True  &           0.062  &           0.025  &       0.055  &       0.013  &      -0.034  &          0.030  \\ 
         Antanas Kascenas  &    Python  &  False  &       True  &           0.495  &           0.463  &       0.406  &       0.397  &      -0.195  &          0.323  \\ 
                     ARUL  &    Python  &  False  &       True  &           0.161  &           0.110  &       0.139  &       0.094  &       0.182  &          0.131  \\ 
           AvivInnovation  &         R  &  False  &       True  &           0.232  &           0.221  &       0.249  &       0.210  &      -0.012  &          0.202  \\ 
                 BRIC\_LB  &    MATLAB  &  False  &       True  &           0.455  &           0.415  &       0.376  &       0.338  &      -0.481  &          0.250  \\ 
Can I get your signature?  &    Python  &  False  &       True  &           0.541  &           0.524  &       0.433  &       0.434  &      -0.123  &          0.360  \\ 
                       CH  &    Python  &  False  &      False  &           0.105  &          -0.283  &       0.104  &      -0.350  &      -0.527  &         -0.121  \\ 
                    CIBIM  &    Python  &  False  &       True  &           0.357  &           0.345  &       0.318  &       0.324  &      -0.201  &          0.251  \\ 
       cinc\_sepsis\_pass  &    Python  &  False  &      False  &           0.220  &          -0.059  &       0.142  &      -0.086  &      -0.665  &         -0.036  \\ 
                  CIS2216  &    Python  &  False  &       True  &           0.176  &           0.148  &       0.156  &       0.136  &       0.193  &          0.155  \\ 
                  claguet  &    Python  &  False  &      False  &           0.000  &           0.000  &       0.000  &       0.000  &       0.000  &          0.000  \\ 
         CQUPT\_Just\_Try  &    Python  &  False  &       True  &           0.674  &           0.660  &       0.392  &       0.381  &      -0.174  &          0.313  \\ 
                 CTL-Team  &    Python  &  False  &       True  &           0.489  &           0.469  &       0.401  &       0.407  &      -0.094  &          0.337  \\ 
               Doctor Who  &    Python  &   True  &       True  &           0.498  &           0.437  &       0.356  &       0.320  &      -0.285  &          0.259  \\ 
                 ECGuru10  &    MATLAB  &   True  &      False  &           0.406  &            None  &       0.372  &       0.358  &      -0.280  &          0.281  \\ 
                      ESL  &    MATLAB  &  False  &       True  &           0.799  &           0.781  &       0.314  &       0.296  &      -0.163  &          0.245  \\ 
             FlyingBubble  &    Python  &  False  &       True  &           0.642  &           0.608  &       0.420  &       0.401  &      -0.156  &          0.337  \\ 
                    haola  &    Python  &  False  &      False  &            None  &            None  &       0.214  &       0.254  &      -0.329  &          0.154  \\ 
                     IADI  &    Python  &   True  &       True  &           0.446  &           0.401  &       0.387  &       0.365  &      -0.148  &          0.309  \\ 
                    IMSAT  &    MATLAB  &  False  &       True  &           0.307  &           0.243  &       0.274  &       0.231  &      -0.261  &          0.190  \\ 
              Infolab USC  &    Python  &   True  &       True  &           0.537  &           0.481  &       0.378  &       0.347  &      -0.262  &          0.284  \\ 
                  ISIBrno  &    Python  &  False  &       True  &           0.409  &           0.357  &       0.361  &       0.327  &      -0.182  &          0.278  \\ 
            Kent Ridge AI  &    Python  &  False  &      False  &          -0.050  &          -0.282  &      -0.047  &      -0.288  &      -0.361  &         -0.164  \\ 
                    Kriss  &    Python  &  False  &      False  &            None  &           0.209  &       0.251  &       0.169  &      -0.117  &          0.177  \\ 
                     LDBR  &    Python  &  False  &       True  &           0.217  &           0.202  &       0.199  &       0.193  &       0.062  &          0.179  \\ 
            Leicester Fox  &    Python  &  False  &       True  &           0.821  &           0.783  &       0.288  &       0.239  &       0.014  &          0.237  \\ 
                   njuedu  &    Python  &  False  &       True  &           0.416  &           0.276  &       0.401  &       0.278  &      -0.207  &          0.282  \\ 
                   NN-MIH  &    Python  &  False  &       True  &           0.460  &           0.422  &       0.414  &       0.373  &      -0.174  &          0.323  \\ 
            OneMoreSecond  &    Python  &  False  &       True  &           0.334  &           0.255  &       0.305  &       0.223  &      -0.332  &          0.195  \\ 
        PhysioNet Example  &    Python  &  False  &       True  &           0.241  &           0.096  &       0.220  &       0.099  &       0.044  &          0.159  \\ 
Ping An Health Technology  &    Python  &  False  &       True  &           0.628  &           0.601  &       0.414  &       0.400  &      -0.182  &          0.331  \\ 
                     PIPI  &    Python  &   True  &      False  &           0.000  &           0.000  &       0.000  &       0.000  &       0.000  &          0.000  \\ 
                PKU\_DLIB  &    Python  &  False  &       True  &           0.439  &           0.400  &       0.402  &       0.386  &      -0.169  &          0.321  \\ 
                     PLUX  &    Python  &  False  &       True  &           0.239  &           0.229  &       0.206  &       0.214  &       0.055  &          0.188  \\ 
                    pqlab  &    Python  &  False  &       True  &           0.773  &           0.714  &       0.287  &       0.260  &      -0.073  &          0.231  \\ 
                     prna  &    Python  &  False  &       True  &           0.511  &           0.474  &       0.411  &       0.389  &      -0.159  &          0.328  \\ 
        Purdue University  &    Python  &  False  &       True  &           0.213  &           0.092  &       0.161  &       0.062  &      -0.323  &          0.066  \\ 
                       Py  &    Python  &  False  &       True  &           0.756  &           0.763  &       0.022  &       0.021  &       0.046  &          0.025  \\ 
                     QLab  &    Python  &   True  &       True  &           0.423  &           0.366  &       0.342  &       0.324  &      -0.161  &          0.270  \\ 
                 RadAsadi  &    Python  &  False  &       True  &           0.424  &           0.358  &       0.382  &       0.335  &       0.048  &          0.323  \\ 
                  R\&Hope  &    Python  &  False  &       True  &           0.537  &           0.510  &       0.374  &       0.344  &      -0.080  &          0.304  \\ 
                  RoBusto  &    MATLAB  &  False  &      False  &           0.106  &            None  &       0.062  &       0.156  &      -0.518  &          0.014  \\ 
                    Rx-LR  &    Python  &  False  &       True  &           0.277  &           0.231  &       0.247  &       0.198  &      -0.037  &          0.194  \\ 
                SailOcean  &    Python  &  False  &       True  &           0.531  &           0.516  &       0.430  &       0.422  &      -0.048  &          0.364  \\ 
                      SBU  &    Python  &  False  &       True  &           0.507  &           0.472  &       0.408  &       0.402  &      -0.154  &          0.332  \\ 
               ScuDicaLab  &    Python  &  False  &      False  &          -0.005  &          -0.031  &      -0.003  &      -0.032  &      -0.083  &         -0.023  \\ 
               Separatrix  &    Python  &  False  &       True  &           0.680  &           0.649  &       0.422  &       0.395  &      -0.146  &          0.339  \\ 
         Sepsis' debugger  &    Python  &  False  &       True  &           0.018  &           0.139  &       0.015  &       0.004  &      -0.062  &          0.002  \\ 
             SepsisFinder  &    Python  &  False  &       True  &           0.578  &           0.524  &       0.378  &       0.315  &      -0.318  &          0.266  \\ 
         Sepsis ReSepsion  &    MATLAB  &   True  &      False  &           0.271  &            None  &       0.227  &       0.036  &      -0.457  &          0.076  \\ 
                   Sepsyd  &    Python  &  False  &       True  &           0.501  &           0.479  &       0.409  &       0.396  &      -0.042  &          0.345  \\ 
                     SFAA  &    Python  &  False  &       True  &           0.479  &           0.445  &       0.405  &       0.376  &      -0.148  &          0.323  \\ 
              Shivpatidar  &    MATLAB  &  False  &       True  &           0.532  &           0.480  &       0.390  &       0.386  &      -0.212  &          0.309  \\ 
                    SHODH  &    MATLAB  &  False  &       True  &           0.014  &           0.020  &       0.012  &       0.013  &       0.017  &          0.013  \\ 
                s(k)eptic  &    MATLAB  &  False  &      False  &            None  &            None  &       0.371  &       0.343  &      -0.230  &          0.282  \\ 
 SOS: Searching of Sepsis  &    MATLAB  &  False  &       True  &           0.529  &           0.496  &       0.399  &       0.392  &      -0.220  &          0.314  \\ 
                   strawc  &    Python  &  False  &      False  &            None  &            None  &       0.220  &       0.099  &       0.044  &          0.159  \\ 
          Team\_Tesseract  &    MATLAB  &  False  &       True  &           0.307  &           0.242  &       0.274  &       0.233  &      -0.246  &          0.192  \\ 
         Terminator\_CUET  &    Python  &  False  &       True  &           0.395  &           0.370  &       0.349  &       0.337  &      -0.259  &          0.264  \\ 
    The memristive agents  &    MATLAB  &  False  &       True  &           0.303  &           0.253  &       0.270  &       0.236  &      -0.176  &          0.200  \\ 
    The Sepsis Detectives  &    Python  &  False  &      False  &          -0.301  &          -1.107  &      -0.321  &      -1.146  &      -2.307  &         -0.841  \\ 
    The Septic Think Tank  &    MATLAB  &  False  &       True  &           0.434  &           0.425  &       0.372  &       0.378  &      -0.218  &          0.296  \\ 
                   Tricog  &    Python  &  False  &       True  &           0.312  &           0.161  &       0.249  &       0.152  &      -0.327  &          0.142  \\ 
        TU Dresden - IBMT  &    Python  &  False  &       True  &           0.205  &           0.127  &       0.194  &       0.106  &      -0.204  &          0.114  \\ 
                 UAlberta  &         R  &  False  &       True  &           0.428  &           0.381  &       0.396  &       0.375  &      -0.060  &          0.329  \\ 
               UBC - DHIL  &         R  &  False  &       True  &           0.319  &           0.273  &       0.296  &       0.268  &       0.007  &          0.249  \\ 
            UCAS\_Bigbird  &    Python  &  False  &       True  &           0.426  &           0.391  &       0.373  &       0.360  &      -0.179  &          0.295  \\ 
          UCAS\_DataMiner  &    Python  &  False  &       True  &           0.555  &           0.538  &       0.406  &       0.373  &      -0.215  &          0.313  \\ 
                ucas-star  &    Python  &  False  &       True  &           0.334  &           0.242  &       0.313  &       0.252  &       0.003  &          0.253  \\ 
UHN\_rand\_num\_generator  &    Python  &  False  &       True  &           0.436  &           0.354  &       0.370  &       0.316  &      -0.263  &          0.269  \\ 
            UM Antiseptic  &    Python  &  False  &       True  &           0.321  &           0.279  &       0.288  &       0.268  &      -0.212  &          0.215  \\ 
             UND\_BERCLAB  &    Python  &  False  &       True  &           0.214  &           0.039  &       0.172  &      -0.000  &      -0.682  &          0.005  \\ 
          USF-Sepsis-Phys  &    Python  &  False  &       True  &           0.344  &           0.366  &       0.303  &       0.348  &      -0.448  &          0.217  \\ 
                     USST  &    Python  &  False  &       True  &           0.343  &           0.267  &       0.282  &       0.200  &      -0.652  &          0.133  \\ 
                 UVA CAMA  &    MATLAB  &  False  &       True  &           0.435  &           0.314  &       0.402  &       0.318  &      -0.145  &          0.303  \\ 
                     VGTU  &    MATLAB  &  False  &       True  &           0.700  &           0.722  &       0.036  &       0.013  &      -0.078  &          0.014  \\ 
                       vn  &    Python  &  False  &       True  &           0.419  &           0.350  &       0.387  &       0.351  &      -0.251  &          0.291  \\ 
      Whitaker's Warriors  &    Python  &   True  &       True  &           0.044  &           0.012  &       0.040  &       0.006  &      -0.016  &          0.022  \\ 
                  WIN-UAB  &    Python  &  False  &       True  &           0.374  &           0.293  &       0.344  &       0.267  &      -0.247  &          0.241  \\ 
                      WML  &    Python  &  False  &      False  &            None  &           0.370  &       0.191  &       0.289  &      -0.241  &          0.164  \\ 
               XLS-IMECAS  &    MATLAB  &  False  &      False  &            None  &            None  &       0.255  &       0.184  &      -0.307  &          0.158  \\ 
            Yuanfang Guan  &    Python  &  False  &       True  &           0.570  &           0.527  &       0.422  &       0.410  &      -0.166  &          0.340  \\ 
                  ywangda  &    Python  &   True  &      False  &            None  &           0.294  &       0.327  &       0.267  &      -0.243  &          0.233\\\hline
\caption{Table of the 88 algorithms that were successfully scored on the full test set: team name, choice of programming language, GPU use, focus of this analysis (successful on each training and test set and a positive utility score on each training set), and the clinical utility score on each training and test set.}
\label{table:teams}
\end{longtable}}

\end{document}